\documentclass[letterpaper]{article} 
\usepackage[pdftex]{graphicx} 
\usepackage{lineno}
\usepackage{amsmath}
\usepackage{hyperref} 

\usepackage[left=0.98in, right=0.98in, top=1.0in, bottom=1.0in]{geometry}

\newcommand\beq{\begin{equation}}
\newcommand\eeq{\end{equation}}

\newcommand\bmat{\begin{bmatrix}}
\newcommand\emat{\end{bmatrix}}

\begin{document}
\setpagewiselinenumbers        
\modulolinenumbers[1]          


\title{ Behavior Trees as a Representation for Medical Procedures}

\author{
        Blake Hannaford\\
        Biorobotics Laboratory\\
        Department of Electrical Engineering \\
        The University of Washington
}

\date{\today}

\maketitle

\paragraph{Abstract}
Behavior trees (BTs) emerged from video game development as a graphical language for modeling
intelligent agent behavior.  BTs have several properties which are attractive for modeling 
medical procedures including human-readability, authoring tools, and composability.  This paper 
will illustrate construction of BTs for exemplary medical procedures
\footnote{We are pleased to acknowledge support from National Science Foundation grant \#IIS-1637444 and collaborations on that
project with Johns Hopkins University and Worcester Polytechnic Institute.}
.

\section{Introduction}
Prior to about 2010, the term “Behavior Tree” (BT) was used idiosyncratically by several authors, but around that time a literature began to emerge around a tree model of behaviors used by the game industry for AI-based characters\cite{halo,lim2010evolving}.  These BTs assume that units of intelligent behavior (such as sensing procedures or sense/action pairs) can be programmed such that they perform a piece of an overall task/behavior, and that they can determine and return a 1-bit result indicating success or failure.  These units are the leaves of BTs.   The level of abstraction of BT leaves is not specified by the BT formalism and varies from one application to another. In the context of medical robotics, they could be things such as a guarded move, a precision cutting action, acquisition of an ultrasound image, creation of a plan, etc. Earlier medical robotics systems such as Robodoc addressed the problem with, for example, scripting languages\cite{kazanzides1992architecture}.
Recent literature has applied BTs to UAV control\cite{ogren2012increasing}, humanoid robotic control\cite{tumova2014maximally}, and human-robot cooperation in manufacturing\cite{paxton2017costar}.
Theoretical classification of BTs has been studied by several authors\cite{TRO17Colledanchise} which has formally related BTs to Finite State Machines (FSMs). BTs have advantages of modularity and scalability with respect to finite state machines.  Other theoretical studies have related BTs to Hybrid Dynamical Systems\cite{colledanchise2014behavior}, humanoid robotic behavior\cite{tumova2014maximally}, and derivation of correctness guarantees\cite{colledanchise2017synthesis}. 
Software packages and ROS implementations\footnote{\url{https://github.com/miccol/ROS-Behavior-Tree}} are now available\cite{marzinotto2014towards}.  
Several of the above references
have ample introductory material and examples of BT concepts. 

When implementing intelligent behavior with BTs, 
the designer of a robotic control system breaks the task down into modules (BT leaves) 
which return either ``success" or ``failure" when called by parent nodes.  All higher level nodes 
define composition rules to combine the leaves including: Sequence, Selector, and Parallel node types. 
A Sequence node defines the order of execution of leaves and returns success if all leaves succeed in order. 
A Selector node (also called ``Priority" node by some authors) tries leaf behaviors in a fixed order, 
returning success when a node succeeds, and failure if all leaves fail.
Decorator nodes have a single child and can modify behavior of their children with rules such as ``repeat 
until $X>0$".
BTs have been explored in the context of humanoid
robot control
\cite{marzinotto2014towards,colledanchise2014performance,bagnell2012integrated} and as a modeling language for intelligent robotic surgical procedures \cite{hu2015semi}.

In this paper, we explore the use of BTs to represent medical procedures (often referred to as algorithms). 
We will illustrate this use by converting published algorithms given in the literature to BTs.  

\newpage

\section{Example 1. Blood Draw}

\begin{figure}\centering
\includegraphics[width=4.750in]{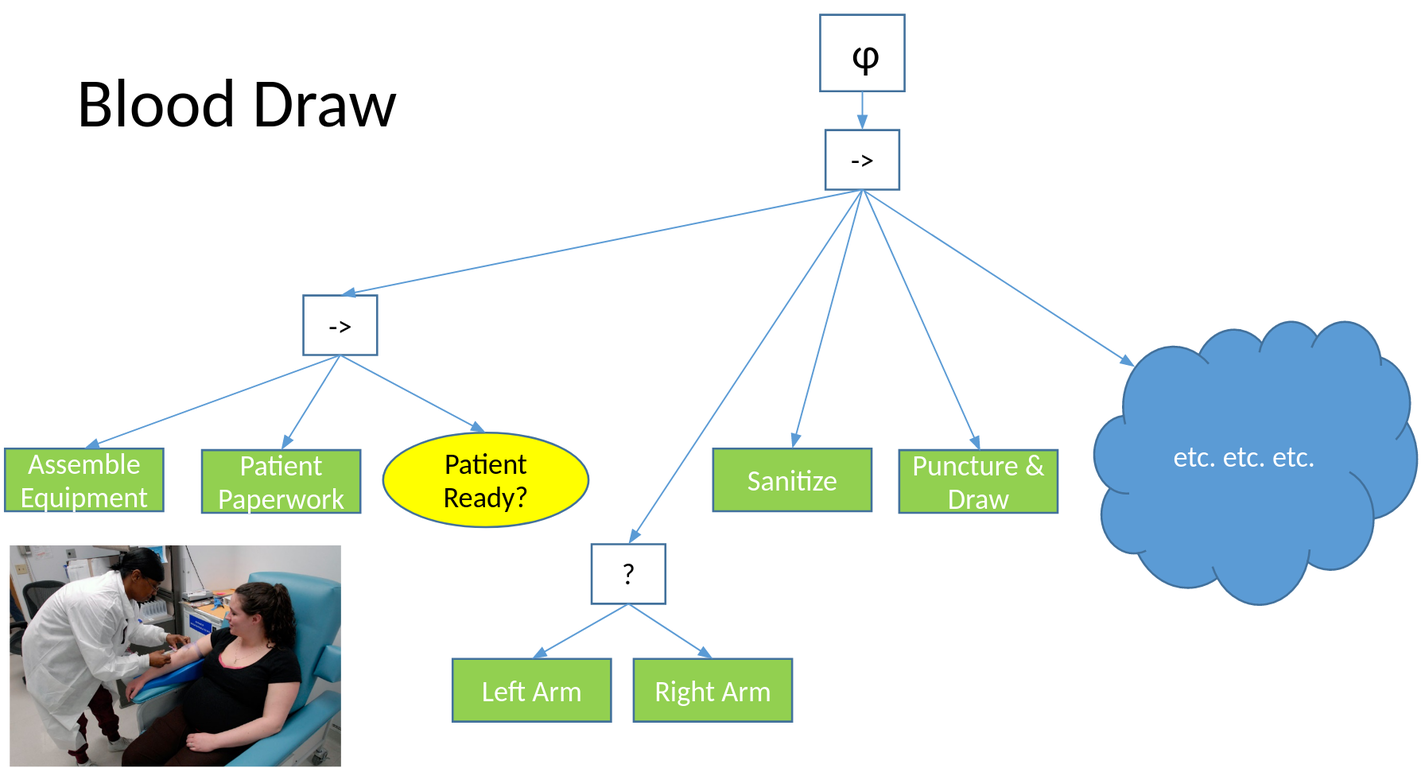}
\caption{BT constructed for the emergency airway procedure of \cite{stephens2009success}.}\label{BloodDrawBTfig}
\end{figure}

The World Health Organization issues a 
\footnote{\url{http://who.int/infection-prevention/tools/injections/drawing_blood_best/en/}}{best practices document}
on drawing blood for medical tests (phlebotomy)\cite{WHO_bestpractices}. 
This over-100-page document gives many details for each step of what is 
mostly a serial process with few branches.  A BT representing 
the first several steps of this process was developed and is represented 
in Figure \ref{BloodDrawBTfig}.

The root (top) node, $\Phi$, encapsulates task start, task end, and task ``succeed/fail" status.  Its only child (BT root always has just one child) 
is a Sequence node ($\to$) indicating that execution will be passed to 
each child in sequence from left to right as shown, with ``Failure" returned
by the node if any child returns ``Failure".   The first child is also
a sequence node which secures equipment and paperwork, and assesses
the overall readiness of the patient. In this and subsequent diagrams, 
we use Yellow to indicate a query or sensing operation which returns 
``Success" or ``Failure" based only on sensing of the world state (in this
case if the patient is ready).  Green leafs indicate tasks that are 
physically performed.   The second child of the main ``Sequence" node
is a ``Selector" node in which the phlebotomist determines whether 
or not a suitable vein is present in the left or right arm.   If neither
arm shows a suitable vein then the ``Selector" node will fail and 
that failure will propagate up to the Sequence and in turn to the tree itself.  

\newpage
\section{Example 2. Emergency Airway Ventilation}

\begin{figure}[h]
\vspace{1.0in}
\href{http://www.sciencedirect.com/science/article/pii/S0952818004000480}{[Left]}
\href{http://www.tandfonline.com/doi/full/10.1080/10903120601023370}{[RIGHT]}
(Figures pending Copyright permissions)
\vspace{1.0in}
\caption{Existing representations of airway establishment include Left: American Society of Anesthesiology\cite{rosenblatt2004airway}, 
Right: Davis et al. \cite{davis2007effectiveness} }\label{airwaytradfigs}
\end{figure}

Human life will expire in minutes if the upper airway is blocked.   
A medical team thus must quickly follow 
a best practice sequence of interventions until airflow is reestablished. 
Restoration of airway consists of a rapid succession of increasingly 
invasive steps, starting with insertion of a laryngoscope, and, as a last
resort, surgical opening of the airway through crychothyroidotomoy. 
The literature on airway restoration algorithms contains many diagrammatic languages for 
representation of the airway algorithm.   One such diagram
includes an exception in the form of a box to the side of a flowchart containing:

\begin{quotation} 
        ``If SpO2 drops to 93\% at any point:
        Facemask + OPA or SGA. If no ETCO2 with best attempts, 
        progress to surgical airway."\cite{stephens2009success}
\end{quotation}

This box can is explicitly outside the flowchart but indicates a concurrent 
monitoring and interrupt task which is hard to represent in the original selected notation. 

\begin{figure}[h]\centering
\includegraphics[width=4.750in]{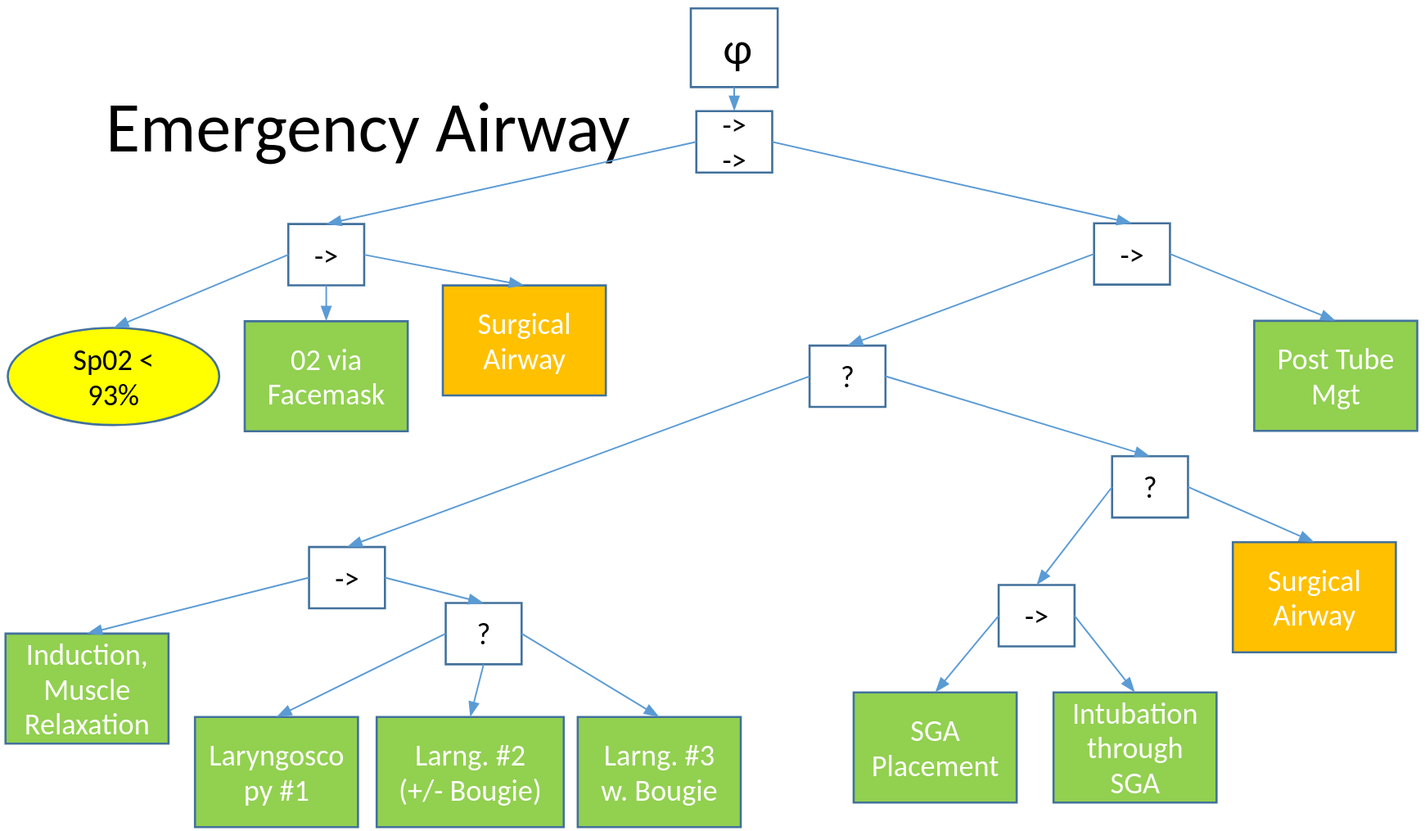}
\caption{BT constructed for the emergency airway procedure of \cite{stephens2009success}.}\label{airwayBTfig}
\end{figure}

We constructed a BT for the airway procedure (Figure \ref{airwayBTfig}) based on \cite{stephens2009success} and interpreted by
\url{https://emcrit.org/racc/shock-trauma-center-failed-airway-algorithm/}.
 The first logic node (directly below $\Phi$) is a
``parallel" node, which indicates that its children should execute concurrently.  The left-most child of the
parallel node represents the concurrent
monitoring procedure represented as a side box in \cite{stephens2009success}.
The right branch, defining the main algorithm,  contains a sequence node ($\to$).  Its left-most child in
turn is a ``Selector" node which allows for alternative methods, returning when the first of its children
succeeds. It can be verified that in the   procedure depicted by this BT, the surgical airway procedure (as seen in the movies) 
is a last-resort which only is attempted when laryngoscopy (up to 3 attempts) and Intubating SGA placement (two attempts) fail.

Compared to the flowchart of \cite{stephens2009success}, the BT is a uniform representation which clearly labels alternative strategies and fallbacks (via the ``?'' (Selector) nodes), and is amenable to direct software execution (assuming code modules (such as for example ROS nodes) are available for each
leaf.

\newpage
\section{Example 3. Simulated Tumor Margin Ablation}

\begin{figure}\centering
\includegraphics[width=4.50in]{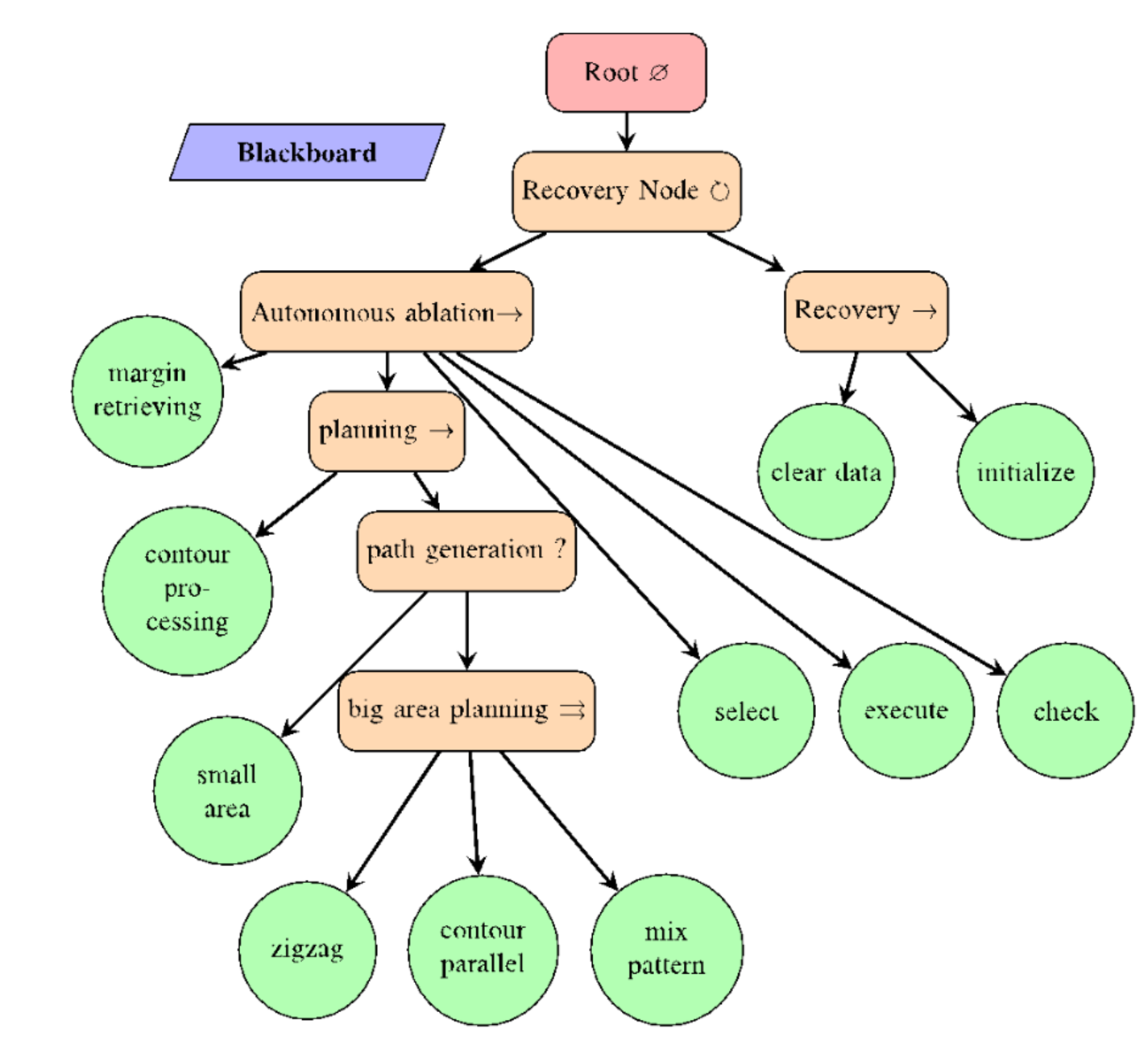}
\caption{BT constructed detection and ablation/treatment of positive 
tumor margins\cite{hu2015semi}. A blackboard data store is commonly 
used with BTs to allow them to share information.}\label{HuTumorAblationBT}
\end{figure}

In recent bench-top surgical robotics experiments\cite{hu2015semi,hu2017semi}
a system was developed which illustrated a future surgical scenario for 
treatment of glioma.  In this scenario, a surgeon will expose the tumor
and manually remove it, but the problem remains of detecting and treating
any remaining tumor material at the edge of the resulting cavity. 
In many cancer surgeries, a margin of up to a centimeter is taken around 
the tumor to increase the odds that no residual cells are left behind.

In this work, Hu et al. assumed the existence of a currently-under-development 
biomarker for brain tumors\cite{veiseh2007tumor} which would allow 
residual tumor material to be detected through fluorescence.  They 
developed a robotic system which could scan the cavity for simulated fluorescence, 
detect a response, and plan and execute one or more treatment plans. 

The BT we developed (Figure \ref{HuTumorAblationBT}) performs this task, 
and checks up to four planning algorithms (lower left leaves) for appropriateness depending on the area and
shape of the detected fluorescent region.  Notably Hu et al., developed a new type of node, the ``Recovery"
node, which is able to fall back to a recovery tree in the event of a task failure.  

Another notable feature of this Medical BT is the ``Select" leaf.   In this 
implementation, selecting of the plan from among several computed plans, 
was performed by manual input from a surgeon.   Thus the BT framework can 
easily incorporate manual steps into a complex and composable procedure.  
Furthermore, should an automated function be developed with sufficient 
confidence, it can easily be dropped in to the select leaf node of the BT.

\newpage
\section{Conclusion}
The use of BTs for medical algorithms is still conceptual.  Anticipated uses to be developed and validated in the future include
\begin{itemize}
\item Documentation of ``standard of care'' algorithms for human medical providers.
\item Execution frameworks for automated medical robotic tasks
\item Description and coordination of 
Human-Robot-Collaborative Systems\cite{kragic2005human,paxton2017costar}
in medical robotics.
\end{itemize}
Compared to Finite State Machines, Hidden Markov Models, and similar
approaches, BTs afford a human-readable and writable representation through
its small number of relatively easy to understand combinatorial operators: ``Sequence'' and ``Selector'', 
and the ease by which BTs can be combined (using those same operators).
These properties seem to be well matched to 
conventional human thinking about procedures. 

There are also limitations of BTs which need further exploration and
elucidation to make sure they are used appropriately.  For example
\begin{itemize}
\item BTs do not have an explicit ``interrupt'' mechanism by which an 
ongoing procedure can be stopped. 
\item New safety checking mechanisms (such as the ``Recovery'' node described in 
Hu et. al.\cite{hu2015semi,hu2017semi})  need further development and
unification. 
\item Learning of BTs is still very much an open problem.   Initial study\cite{lim2010evolving}
and more recent works\cite{colledanchise2015learning,arXivHannafordHZL16} suggest 
some possibilities for on-line autonomous performance improvement. 
\end{itemize}

\newpage
\bibliography{adapt_BTs_icra,Proposal}
\end{document}